%% file: paper_naacl21.tex
\pdfoutput=1

\documentclass[11pt]{article}

\usepackage[]{acl}

\usepackage{times}
\usepackage{latexsym}
\usepackage{booktabs}
\usepackage{graphicx}
\usepackage{soul}
\usepackage{xcolor}

\usepackage[colorinlistoftodos,prependcaption,textsize=tiny]{todonotes}

\newcounter{sm}
\usepackage[T1]{fontenc}

\usepackage[utf8]{inputenc}

\usepackage{microtype}

\title{Counterfactual Data Augmentation improves  \\ Factuality of Abstractive Summarization}

\author{Dheeraj Rajagopal, Siamak Shakeri$^\dagger$, Cicero Nogueira dos Santos$^\dagger$ \\ 
\textbf{Eduard Hovy}, \textbf{Chung-Ching Chang$^\dagger$} \\
  Language Technologies Institute, Carnegie Mellon University \\
  $^\dagger$ Google Research \\
  \texttt{\{dheeraj,hovy\}@cs.cmu.edu} \\ \texttt{\{siamaks,cicerons,ccchang\}@google.com} \\} 

\begin{document}
\maketitle

\input{custom_commands}
\begin{abstract}
Abstractive summarization systems based on pretrained language models often generate coherent but factually inconsistent sentences. In this paper, we present a counterfactual data augmentation approach where we augment data with perturbed summaries that increase the training data diversity. Specifically, we present two augmentation approaches based on replacing (i) entities from other and the same category and (ii) nouns with their corresponding WordNet hypernyms. We show that augmenting the training data with our approach improves the factual correctness of summaries without significantly affecting the ROUGE score. We show that in two commonly used summarization datasets (\cnn and \xsum), we improve the factual correctness by about 2.5 points on average \footnote{All the code and data will be released soon}. 
\end{abstract}

\input{sections/introduction}
\input{sections/approach}

\input{sections/experiments}

\input{sections/related_work}
\input{sections/conclusion}

\bibliography{custom}
\bibliographystyle{acl_natbib}

\end{document}

%% file: custom_commands.tex
\newcommand{\dheeraj}[1]{\textcolor{blue}{[#1 \textsc{--dheeraj}]}}
\newcommand{\ccchang}[1]{\textcolor{orange}{[#1 \textsc{--ccchang}]}}
\newcommand{\siamak}[1]{\textcolor{green}{[#1 \textsc{--siamak}]}}
\newcommand{\cicero}[1]{\textcolor{green}{[#1 \textsc{--cicero}]}}

\newcommand{\cnn}{CNN/DailyMail~}
\newcommand{\xsum}{XSum~}
\newcommand{\metric}{E2E NLI~}
\newcommand{\tfive}{\textsc{T5-LARGE}~}
\newcommand{\wn}{\textsc{WN-Hyper}~}
\newcommand{\gb}{\textsc{ER}~}
\newcommand{\cat}{\textsc{Cat-ER}~}
\newcommand{\rouge}{\textsc{ROUGE}~}

\newcommand{\squishlist}{
  \begin{list}{$\bullet$}
    { \setlength{\itemsep}{0pt}      \setlength{\parsep}{3pt}
      \setlength{\topsep}{3pt}       \setlength{\partopsep}{0pt}
      \setlength{\leftmargin}{1.5em} \setlength{\labelwidth}{1em}
      \setlength{\labelsep}{0.5em} } }
\newcommand{\reallysquishlist}{
  \begin{list}{$\bullet$}
    { \setlength{\itemsep}{0.1pt}    \setlength{\parsep}{0.1pt}
      \setlength{\topsep}{0.5pt}     \setlength{\partopsep}{0.5pt}
      \setlength{\leftmargin}{0.5em} \setlength{\labelwidth}{0.5em}
      \setlength{\labelsep}{0.2em} } }

 \newcommand{\squishend}{
     \end{list} 
 }

%% file: sections/introduction.tex
\section{Introduction} 

Hallucinations or generating facts that can't be substantiated with evidence from the input is still an uphill challenge for generative models \citep{Cao_Wei_Li_Li_2018, falke-etal-2019-ranking, goyal-durrett-2021-annotating}. Standard maximum likelihood training (MLE) has been shown as the reason that models produce inconsistent summaries, making them prone to hallucinations \citep{cao-etal-2020-factual,nan-etal-2021-improving,cao-wang-2021-cliff}. 
In this work, we propose a counterfactual data augmentation approach to improve factuality in summarization without the need for modifying the MLE objective. We show this approach leads to improvements in both factual correctness and increased specificity of the summaries. 

\begin{table}[t]
\begin{tabular}{|p{7cm}|}
\toprule
\textbf{Input :}  ...He had not really spoken about his injuries, she said, although he was aware of the extend of them. `` I don't think he's ready", she said....    \\ \midrule
\textbf{Reference :}  A bus driver was seriously hurt when a steam engine fell off a lorry into the path of his vehicle remains {\color{red}\hl{really poorly}}, his wife has said.   \\ \midrule
\textbf{No Augmentation :}  A bus driver who was seriously injured when he was hit by a steam engine is {\color{red}\hl{making good progress}}, his wife has said.   \\ \midrule
\textbf{Our Work :}  "A bus driver who suffered serious injuries when he was hit by a steam engine is {\color{red}\hl{``not ready" to return to work}}, his wife has said ."   \\ \bottomrule
\end{tabular}
\caption{Sample input document and output summary from our baseline and our approach. Despite showing comparable \rouge scores, we see that the more factually correct summary is produced by our approach.}
\label{tab:intro-example}
\end{table}

\begin{figure*}[th]
\centering
  \includegraphics[width=\linewidth,keepaspectratio]{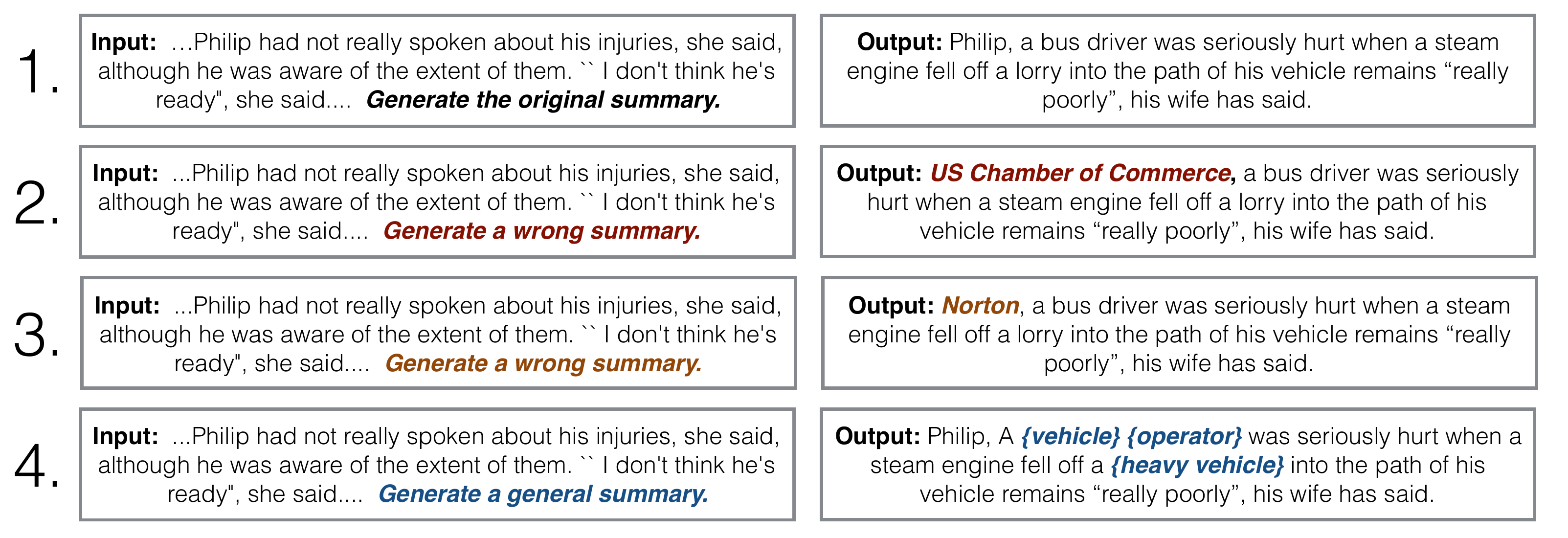}
\caption{Overview of our data augmentation approach. (1) denotes the original dataset while (2), (3), and (4) denote different augmentation strategies. (2) replaces entity in summary with a random entity (3) replaces entity in summary with an erroneous entity of the same category (in this example, another name) and in (4) we replace randomly selected nouns from the summary and replace it with their corresponding WordNet hypernyms. 
The control code $c$ for each transformation are emphasized.}
\label{fig:approach}
\end{figure*}

Data Augmentation approaches have been often used in summarization to improve the diversity of the training samples \citep{fabbri-2020,parida-motlicek-2019-abstract}. Augmentation approaches that do not collect more data often do so by \emph{introducing errors} via entities \citep{nan-etal-2021-improving} or spans of text \citep{cao-wang-2021-cliff} and modifying the original MLE objective. In this work, we explore (i) three augmentation approaches for improved factual consistency and (ii) whether we can improve factuality without modifying the original MLE objective. 


We present the following augmentation strategies where we replace (i) entities with random entities and entities with random entities from the same category and (ii) nouns with their corresponding hypernyms from an ontology. We transform each of the above transformation by transforming the input via a control code (\S\ref{sec:approach}). Figure \ref{fig:approach} shows an example of our overall approach. In contrast to the findings from literature  \cite{nan-etal-2021-improving}, we show that we do not need to modify the original MLE objective, which is commonly cited as the reason for increased hallucinations. 


Our empirical results show that inducing diversity by proposed augmentation improves the factual correctness of summaries without significant change in ROUGE score. In our experiments in two commonly used summarization datasets \cnn and \xsum using the \tfive model, our approach outperforms the original model that is trained only to generate summaries for factual consistency. We show an average of 
$\sim$3 points improvement in both the \cnn and $\sim$2 points improvement in \xsum  with the  recently proposed strong \metric metric for factuality \cite{honovich-etal-2021-q2}. In summary, our results show that counterfactual data augmentation via inducing entity errors and replacing nouns with hypernyms lead to improved factuality. 

%% file: sections/approach.tex
\section{Approach}
\label{sec:approach}

Data augmentation approaches for learning with augmentation \citep{8953916} often rely on transforming the input such that the model is introduced to training data variations during learning. In such an approach, a transformation $t$ is applied to an input sample $\mathbf{x}$ (with a corresponding output $\mathbf{y}$) to get a modified sample $\tilde{\mathbf{x}} = t(\mathbf{x})$. For our approach, we propose a simple transformation where we augment a \emph{control code} $c$ to $\mathbf{x}$ and correspondingly augment the output $\tilde{\mathbf{y}}$ in accordance with the control code as shown in Figure \ref{fig:approach}. For the summarization task, we formulate this is as a learning task where the model learns to optimize the following loss function: 

$$ \mathcal{L}_S = \frac{1}{M} \sum_{j=1}^M \mathcal{L}_{CE} (\mathbf{x}_j, \mathbf{y}_j) +  \mathcal{L}_{CE} (\tilde{\mathbf{x}}_j, \tilde{\mathbf{y}}_j)  $$
where $\mathcal{L}_{CE}$ stands for the cross-entropy loss and $(\mathbf{x}_j, \mathbf{y}_j)$ an input document, summary pair and $(\tilde{\mathbf{x}}_j, \tilde{\mathbf{y}}_j)$ a transformed document, summary pair for $M$ samples. 

\begin{table*}[]
\centering
\begin{tabular}{@{}llllllll@{}}
\toprule
Dataset & Augmentation & \metric & ROUGE-1 & ROUGE-2 & ROUGE-L &  $\Delta_{\texttt{R-L}}$ &  $\Delta_{\texttt{NLI}}$ \\ \midrule
\cnn     & Original     & 61.0   & 42.5    & 19.6    & 39.8    &      -          &  -  \\
        & \gb           & 62.3   & 41.3    & 19.3    & 38.8    &     -1.0         & +1.3 \\
        & \cat          & 63.9   & 41.9    & 19.6    & 38.3    &     -1.5       & +2.9 \\
        & \wn           & 64.2   & 41.1    & 19.3    & 38.8    &     -1.0         & +3.2 \\ \midrule
\xsum    & Original$^*$     & 10.2   & 42.0    & 19.8    & 34.6    &     -         &  - \\
        & \gb           & 10.8   & 40.8    & 18.9    & 33.7    &     -0.9         & +0.6 \\
        & \cat          & 11.9   & 40.8    & 18.9    & 33.7    &    -0.9          & +1.7 \\
        & \wn           & 11.2   & 41.6    & 20.1    & 34.3    &    -0.3          & +1.0 \\ \bottomrule
\end{tabular}
\caption{Results for different augmentation approaches compared to the \emph{Original} model. Note that all models are trained for same number of steps and with 5 random seeds (reporting the average). $*$- denotes our reproduction of the results with the preprocessed dataset as described in \S\ref{subsec:datasets}. $\Delta_{\texttt{R-L}}$ and $\Delta_{\texttt{NLI}}$ refer to difference in the ROUGE-L and E2E NLI scores of the augmented models versus the original, accordingly. Note: Each row specifies individual augmentation.}
\label{tab:main-results}
\end{table*}

\subsection{Data Augmentation}

Factuality in summarization comprises of a diverse range of inconsistencies as described in \citet{pagnoni-etal-2021-understanding}. The prominent errors among them are \emph{entity errors} that captures errors where wrong entities are used as primary arguments and \emph{circumstance errors} where arguments are mistakenly generated. Inspired by this, we propose the following three transformations for our approach to tackle such errors:

\paragraph{Entity Replacement (\gb): } In this perturbation, we replace entities in a reference summary with another random entity in the training dataset that is not present in the current input sample. An example is shown in  Figure \ref{fig:approach} (2), where a \texttt{PERSON} entity \textit{Philip} is replaced by a random entity \textit{US Chamber of Commerce}, an \texttt{ORGANIZATION}. Each of the erroneous sample was augmented with a control code \emph{``generate a wrong summary"} to the input document. 

\paragraph{Categorical Entity Replacement (\cat):} Inspired from the previous entity swap, we replace entities with other entities from similar category. An example is shown in Figure \ref{fig:approach} (3) where a \texttt{PERSON} entity \textit{Philip} is replaced by \textit{Norton}, another entity of the same \texttt{PERSON} category. The categorical entity replacement also uses the same control code as entity replacement due to their similar nature of perturbation. 

\paragraph{WordNet Hypernym Replacement (\wn) :} We also try an augmentation strategy that does not rely on introducing errors in reference summaries. In this approach, we replace a fraction of nouns in each reference summary $\mathbf{y}$ with their corresponding lexical category from an ontology. For this, we rely on WordNet \citep{Fellbaum2000WordNetA} hypernym for noun replacement. Correspondingly, we use the control code \emph{``generate a general summary"} as shown in Figure \ref{fig:approach} (4), where  a nouns such as \textit{bus, driver, lorry} as replaced by their corresponding lexical categories - namely \emph{vehicle, operator, heavy vehicle}. We hypothesize that this transformation provides additional semantics to arguments (predominantly nouns) during generation.

%% file: sections/experiments.tex
\section{Experiments}

With the training data, consisting of $(\mathbf{x}, \mathbf{y})$ and augmented $(\tilde{\mathbf{x}}, \tilde{\mathbf{y}})$ samples, we minimize the loss $\mathcal{L}_S$. For all our experiments, we use the transformer based \tfive \citep{raffelT5} as our base model with learning rate of \texttt{1e-03}, batch size of 32, a dropout of 0.1 with the adafactor optimizer.

\paragraph{Metrics :} 
We use the standard \rouge metric \citep{lin2004rouge} to compare against the baselines. To measure the factual correctness, we adopt the \metric metric proposed by \citet{honovich-etal-2021-q2}. In this metric, we use an NLI model trained on SNLI \citep{bowman-etal-2015-large} to assess whether each sentence of the summary can be inferred from the input document. Towards this, we use the RoBERTa \citep{liu2019roberta} model trained on the SNLI corpus. For each predicted summary $\mathbf{y}_p$, we assign a score of 1 if the NLI model predicts a \emph{entailment} and a score of 0 if the model predicts either \emph{neutral} or \emph{contradiction}. Finally, we average these scores across all the test set to yield an overall factuality score.

\begin{table*}[]
\centering
\resizebox{\textwidth}{!}{%
\begin{tabular}{@{}llll@{}}
\toprule
Reason &  & Original & Augmented \\ \midrule
\begin{tabular}[c]{@{}l@{}}Improvement in specificity: \\ More named entities were mentioned\end{tabular} & 23\% & \begin{tabular}[c]{@{}l@{}}Darlington have appealed to\\ the Football Association \\ over the lack of seating at their stadium.\end{tabular} & \begin{tabular}[c]{@{}l@{}}Darlington have launched \\ an appeal to raise \hl{\pounds 150,000} \\ to buy new seats at their stadium .\end{tabular} \\ \cmidrule(l){3-4} 
 &  & \begin{tabular}[c]{@{}l@{}}UKIP has chosen its candidate \\ for London mayor.\end{tabular} & \begin{tabular}[c]{@{}l@{}}UKIP has chosen its candidate \\ to run for London Mayor \\ \hl{in the 2016 election} \end{tabular} \\ \cmidrule(l){3-4} 
\begin{tabular}[c]{@{}l@{}}Factual correctness improves \\ while ROUGE scores decrease\end{tabular} & 9\% & \begin{tabular}[c]{@{}l@{}}A bus driver who was seriously injured \\ when he was hit by a steam engine \\ is \hl{making good progress}, his wife has said.\end{tabular} & \begin{tabular}[c]{@{}l@{}}A bus driver who suffered serious injuries \\ when he was hit by a steam engine is \hl{"not ready"}\\  to return to work, his wife has said.\end{tabular} \\ \cmidrule(l){3-4} 
 &  & \begin{tabular}[c]{@{}l@{}}Harry Redknapp says he would be happy\\  to be appointed England manager, \\ if Roy Hodgson is \hl{appointed}.\end{tabular} & \begin{tabular}[c]{@{}l@{}}Harry Redknapp says he would be happy to \\ \hl{succeed} Roy Hodgson as England manager .\end{tabular} \\ \bottomrule
\end{tabular}%
}
\caption{Qualitative examples comparing the summaries from the \emph{Original} and augmented dataset models. In our analysis, we found the summaries were similar for about 68\% times. For 23\% of the samples, the improvements in factual correctness came due to increase in specificity with similar \rouge scores. We also highlight 9\% of the samples where the \rouge scores were higher for the non-augmented version despite being factually incorrect. }
\label{tab:qual-examples}
\end{table*}

\subsection{Datasets} 
\label{subsec:datasets}
We select the \cnn and \xsum dataset for our experiments since their factual inconsistency is well documented \citep{kryscinski-etal-2020-evaluating}. 

\paragraph{\cnn :} This dataset proposed by \citet{see-et-al-pointer} is composed of summaries from the news websites CNN and Dailymail. 
\paragraph{\xsum :} The \xsum dataset \citep{narayan-etal-2018-dont} is also a news summarization dataset but targeted towards summaries that are extremely short, thus more challenging\footnote{We use entire dataset for both \cnn and \xsum}. 

We compare our model settings with the variation of the model where the original dataset is not augmented (the \emph{Original} setting). All models were train on 16 TPUv3 slices. 
For fair comparison, baseline and proposed approach were trained for the same number of steps (110000), with 5 random seeds and the results are averaged.  

\paragraph{Preprocessing :} We follow the  preprocessing steps outlined in \citet{Nan2021EntitylevelFC} for the \xsum dataset as following: we remove (i) all sentences from the source document that include only noisy strings such as ``Share this with Email, Facebook, Messenger" and (ii) all sentences in summaries that has an entity that is not present in the source document.  For \cnn, we follow the standard dataset processing without any special preprocessing steps. 

\subsection{Results}

We compare our different augmentation strategies to training the model without augmentation. The results are shown in table \ref{tab:main-results}. The ROUGE scores are comparable to the model that was trained on the original dataset without augmentation. We also observe that augmentation uniformly improves factuality across various augmentation settings. For the \cnn dataset, \wn augmentation performs the best, improving by about 3 points compared to the baseline. We observe a similar trend in the \xsum dataset where augmentation achieves improvement in factuality.




\subsection{Analysis} 

An abstractive summary \cite{rush-etal-2015-neural,gehrmann-etal-2019-generating} paraphrases the input document  copies less fragments of text from the input compared to an extractive summary \cite{Neto2002AutomaticTS, Erkan2004LexRankGL}. While recent state-of-the-art methods for summarization methods focus of abstractive methods, it is shown that as \textit{degree of abstractiveness} is an important indicator of hallucinations in summarization models, and often higher hallucinations correlated with higher \rouge scores for such models \cite{Dreyer2021AnalyzingTA}. 

To understand in-depth the nature of the samples where the factuality improves, we sampled 100 random datapoints from development set of \xsum dataset. Among those samples, we found that about 68\% summaries are similar without significant variance in \rouge and have the same \metric score. From the rest of the samples, we observed the following: (i) \textit{Specificity :} Models trained on augmented data often had more named entity information, in effect making them more specific. Table \ref{tab:qual-examples} show some qualitative examples from the model  
(ii) \textit{Factually Correct with reduced ROUGE :} In about 9\% of the samples, despite being factually correct, resulted in lower \rouge scores compared to the original model. 
In our inspection of these samples, we found that the gold news summaries often omit specific details in the input document and tend to be less \textit{extractive}, and extractive summaries are often more factual than abstractive summaries since they copy higher ratio of phrases from the input document.
We observe the same phenomenon as \citet{Dreyer2021AnalyzingTA} that degree of abstractiveness affects the factuality inversely. 
We believe that this contributed to the lower \rouge scores and improve factuality scores in Table \ref{tab:main-results}. We show some qualitative examples in Table \ref{tab:qual-examples} 

%% file: sections/related_work.tex
\section{Related Work}



To generate factually correct summaries, several approaches have been proposed recently. A line of work focuses on post-correcting summaries after they are generated \citet{dong-etal-2020-multi, cao-etal-2020-factual}. 
\citet{zhu-etal-2021-enhancing} propose adding a graph based layer to existing transformer network to improve factuality by modifying the architecture. 
Contrastive learning approaches \citet{nan-etal-2021-improving,cao-wang-2021-cliff,cao-etal-2020-factual} have also been proposed with data perturbation that introduces additional loss components for improving factuality with the assumption that MLE objective leads to inconsistent summaries.
Our approach proposes a data-augmentation alternative to current approaches but we do not require to modify the loss term during fine-tuning or add special layers to the model. 

%% file: sections/conclusion.tex
\section{Conclusion}

Overall, we propose a counterfactual data augmentation approach to improve factuality in summarization systems. We achieve this using a simple data augmentation approach by perturbing the summaries via inducing errors in entities and replacing nouns with their corresponding WordNet hypernyms. Our results show this simple, yet effective technique, leads to improvements in the factual correctness of the generated summaries.